\documentclass{article}

\pdfoutput=1

\usepackage{algorithm}
\usepackage{algorithmic}

\usepackage{authblk}
\usepackage{amsmath}
\usepackage{amssymb}
\usepackage{graphicx}
\usepackage{subfigure}

\DeclareMathOperator*{\argmin}{arg\,min}

\title{Nonnegative/binary matrix factorization with a D-Wave quantum annealer}
\author[1]{Daniel O'Malley}
\author[1]{Velimir V. Vesselinov}
\author[2]{Boian S. Alexandrov}
\author[3]{Ludmil B. Alexandrov}

\affil[1]{\small Computational Earth Science (EES-16), Los Alamos National Laboratory}
\affil[2]{Physics and Chemistry of Materials (T-1), Los Alamos National Laboratory}
\affil[3]{Theoretical Biology and Biophysics (T-6), Los Alamos National Laboratory}

\date{}

\begin{document}
\maketitle

\begin{abstract}
D-Wave quantum annealers represent a novel computational architecture and have attracted significant interest \cite{johnson2011quantum,ronnow2014defining,gibney2017d}, but have been used for few real-world computations.
Machine learning has been identified as an area where quantum annealing may be useful \cite{servedio2004equivalences,aimeur2006machine,pudenz2013quantum}.
Here, we show that the D-Wave 2X can be effectively used as part of an unsupervised machine learning method.
This method can be used to analyze large datasets.
The D-Wave only limits the number of features that can be extracted from the dataset.
We apply this method to learn the features from a set of facial images.
\end{abstract}

\section{Introduction}
Single-core computational performance relentlessly improved for decades, but recently that progress has begun to slow \cite{danowitz2012cpu}.
As a result, alternative computational architectures have sprung up including multi-core processors \cite{geer2005chip}, graphic processing units \cite{owens2008gpu}, neuromorphic computing \cite{monroe2014neuromorphic}, and application-specific integrated circuits to name a few.
Here we explore the use of another new architecture: quantum annealing \cite{kadowaki1998quantum}.
In particular, we utilize the form of quantum annealing realized with D-Wave hardware \cite{johnson2011quantum,gibney2017d}.
We focus on a machine learning problem based on matrix factorizations, and describe an algorithm for computing these matrix factorizations that leverages D-Wave hardware.
We apply the algorithm to learn features in a set of facial images.

There is an ongoing back-and-forth regarding whether or not D-Wave's hardware provides performance benefits over classical single-core computing \cite{mcgeoch2013experimental,isakov2015optimised,king2017quantum,mandra2017pitfalls}.
Here, we benchmark the performance of two state-of-the-art classical approaches against the performance of the D-Wave.
Much of the debate about the D-Wave's performance has centered on problems that are custom-tailored to fit D-Wave's hardware.
A component of the matrix factorization problems that we study here can be solved on the D-Wave, but is not customized for the D-Wave and represents a real rather than synthetic problem.
The D-Wave outperforms the two classical approaches in a benchmark, but the performance of the two classical approaches suggests a heuristic that could be implemented on a classical computer to outperform the D-Wave in this benchmark.
This mixed result on the performance of the D-Wave compared to classical tools is in agreement with the most recent results \cite{mandra2017pitfalls} showing that, even for custom-tailored problems, the D-Wave does not outperform the best classical heuristics.
Despite this, our results show that the D-Wave can outperform very good classical tools when only a short amount of computational time is allotted to solve these real-world problems.
We also provide a discussion of how future improvements to the algorithm presented here and D-Wave's hardware could improve performance for these matrix factorization problems.

The remainder of this manuscript is organized as follows.
Section \ref{sec:methods} describes the methods used to solve the matrix factorization problems and perform the benchmarking.
Section \ref{sec:results} describes the results we obtained in solving a matrix factorization problem for a set of facial images and the benchmark results.
Finally, section \ref{sec:discussion} discusses the results and indicates how future developments might improve the performance of the D-Wave for this problem.

\section{Methods}\label{sec:methods}
We seek to represent an $n\times m$ matrix, $V$, as the product of two matrices, $W$ and $H$, where $W$ is an $n\times k$ matrix and $H$ is a $k\times m$ matrix.
That is, we wish to find $W$ and $H$ such that
\begin{equation}
  V \approx WH
  \label{eq:abc}
\end{equation}
We impose constraints on $W$ and $H$.
In particular, the components of $W$ must be nonnegative (i.e., $W_{ij} \ge 0$) and the components of $H$ must be binary (i.e., $H_{ij}\in\{0,1\}$).
Since $W$ is a nonnegative matrix and $H$ is a binary matrix, we describe this matrix factorization as Nonnegative/Binary Matrix Factorization (NBMF).
This is in contrast to Nonnegative Matrix Factorization (NMF) \cite{lee1999learning} where $H$ is allowed to take on any nonnegative value, not just 0 or 1.
To satisfy equation \ref{eq:abc}, we utilize an alternating least squares algorithm \cite{lin2007projected} (see Algorithm \ref{alg:als}).

\begin{algorithm}
  \caption{A high-level description of the alternating least squares algorithm that we employ to perform NBMF.}
  \label{alg:als}
  \begin{algorithmic}
	\REQUIRE $V$, $k$
	\ENSURE $W$, $H$
	\STATE Randomly initialize each element of $H$ to be either 0 or 1
	\WHILE {not converged}
	  \STATE W := $\argmin_{X\in {{\mathbb{R}^{+}}^{n\times k}}} ||V - XH||_F + \alpha||X||_F$
	  \STATE H := $\argmin_{X\in\{0,1\}^{k\times m}} ||V - WX||_F$
	\ENDWHILE
  \end{algorithmic}
\end{algorithm}

From algorithm \ref{alg:als}, we focus on utilizing a D-Wave 2X quantum annealer to compute
\begin{equation}
  H = \argmin_{X\in\{0,1\}^{k\times m}} ||V - WX||_F
  \label{eq:bigargmin}
\end{equation}
where $||\cdot||_F$ is the Frobenius norm.
Note that equation \ref{eq:bigargmin} can be solved by solving a set of independent optimization problems (one for each column of $H$).
This is because the variables in the $i^{th}$ column of $H$ impact only the $i^{th}$ column of $WX$, and the variables outside the $i^{th}$ column of $H$ do not impact the $i^{th}$ column of $WX$.
That is, if we denote the $i^{th}$ columns of $H$ and $V$ by $H_i$ and $V_i$, respectively, then
\begin{equation}
  H_i = \argmin_{\mathbf{q}\in\{0,1\}^{k}} ||V_i - W\mathbf{q}||_2
  \label{eq:smallargmin}
\end{equation}
for $i=1,2,\ldots,m$.
This means that we can solve for $H$ by solving a series of linear least squares problems in binary variables.
This type of problem can be readily solved on D-Wave hardware \cite{o2016toq}, as long as the number of binary variables is small.
Equation \ref{eq:bigargmin} involves $km$ binary variables, but equation \ref{eq:smallargmin} involves only $k$ binary variables.
Since the D-Wave 2X imposes severe limitations on the number of binary variables that can be dealt with, this reduction in the number of variables is crucial.
In practice, this means that relatively large datasets can be analyzed (i.e., $n$ and $m$ can be large).
Since the number of variables the D-Wave works with at any given time is determined only by $k$, the D-Wave imposes restrictions only on the number of features (i.e., $k$).

We also compare the performance of the D-Wave to solve equation \ref{eq:smallargmin} with two classical approaches.
One, called qbsolv \cite{dwave2017qbsolv}, utilizes is an efficient, open-source implementation of tabu search \cite{glover1989tabu,glover1990tabu}.
The other utilizes the JuMP \cite{dunning2015jump} modeling language and a state-of-the-art mathematical programming tool called Gurobi \cite{gurobi2017gurobi}.
The performance is compared using a cumulative time-to-targets benchmark that is a variation of the time-to-target benchmark \cite{king2015benchmarking}.
In the course of executing algorithm \ref{alg:als}, equation \ref{eq:smallargmin} must be solved many times for different values of $i$ and $W$.
Each time this equation is solved, the D-Wave is given a fixed number of annealing cycles to minimize $||V_i - W\mathbf{q}||_2$ (we look at examples where the number is fixed at $10$, $10^2$, $10^3$, and $10^4$).
Each annealing cycle results in one approximate solution to equation \ref{eq:smallargmin}, and we denote the best of these approximate solutions by $H_i^*$.
The best solution, is used to generate a target value for the objective function, $||V_i - W H_i^*||_2$, that the classical approaches (qbsolv and Gurobi) must match.
The cumulative time-to-targets benchmark computes the cumulative amount of time it takes for qbsolv or Gurobi to find a solution that is at least as good (in terms of $||V_i - W H_i||_2$) as the best solution found by the D-Wave for each instance of equation \ref{eq:smallargmin} that is encountered in executing algorithm \ref{alg:als}.
This cumulative time-to-targets is then compared with the amount of annealing time used by the D-Wave.
If qbsolv or Gurobi take more than 10 minutes to reach an individual target, the time to reach that target is set to 10 minutes.
The 10 minute limit is an expedient that enables the analysis to be run in a reasonable amount of time.
Gurobi never reached the 10 minute limit, but qbsolv did in a number of cases.

\subsection{Programming the D-Wave}
D-Wave quantum annealers deal natively with quadratic, unconstrained, binary optimization (QUBO) problems \cite{mcgeoch2014adiabatic}.
These problems are associated with objective functions that have the form
\begin{equation}
  f(\mathbf{q}) = \sum_i a_i q_i + \sum_{i<j} b_{ij} q_i q_j
\end{equation}
where $\mathbf{q}=(q_1, q_2, \ldots, q_n)$.
One might think of a $0^{th}$ order approximation of the D-Wave's behavior as being that each anneal returns a vector, $\mathbf{q}$, so that $f(\mathbf{q})$ is minimized.
A $1^{st}$ order approximation of the D-Wave's behavior is that each anneal returns a sample, $\mathbf{q}$, from a Boltzmann distribution where $f(\mathbf{q})$ is the energy.
Both of these approximations are inexact, but highlight the basic behavior of the D-Wave: each annealing cycle returns a sample, $\mathbf{q}$, which tends to make $f(\mathbf{q})$ small.
Equation \ref{eq:smallargmin} can be readily put into the form of a QUBO by setting \cite{o2016toq}
\begin{eqnarray}
	\label{eq:bls_qubo_v}
	a_j &=& \sum_k W_{kj}(W_{kj} - 2V_{ij}) \\
	\label{eq:bls_qubo_w}
	b_{jk} &=& 2 \sum_{l} W_{lj} W_{lk}
\end{eqnarray}
Having reformulated equation \ref{eq:smallargmin} in this way, the quadratic coefficients in the QUBO, $b_{ij}$, are generally all nonzero. 
However, the D-Wave's hardware imposes sparsity constraints on the $b_{ij}$.
These constraints can be overcome via embedding \cite{choi2008minor,choi2011minor}, where multiple physical qubits are used to represent a single binary variable.
Since the $b_{ij}$'s in our problems are never exactly zero, a complete graph with the number of nodes equal to the number of binary variables must be embedded in the graph imposed by the D-Wave 2X chip.
If a D-Wave 2X chip had no defects (i.e., if all the qubits and couplers were available for use), the maximum number of binary variables that could be used for these problems is 49 (if the matrix formed by the elements $b_{ij}$ had some natural sparsity to it, this number would increase).
However, some of the qubits and couplers on these chips are not available for use, and as the number of binary variables increases, the number of physical qubits required to represent each variable increases.
Since using a larger number of physical qubits to represent a single binary variable is associated with poor performance, we limit our study to 35 binary variables.
In this case, we found an embedding where each binary variable is represented by at most 19 physical qubits.

\section{Results}\label{sec:results}
\begin{figure}
  \includegraphics[width=\linewidth]{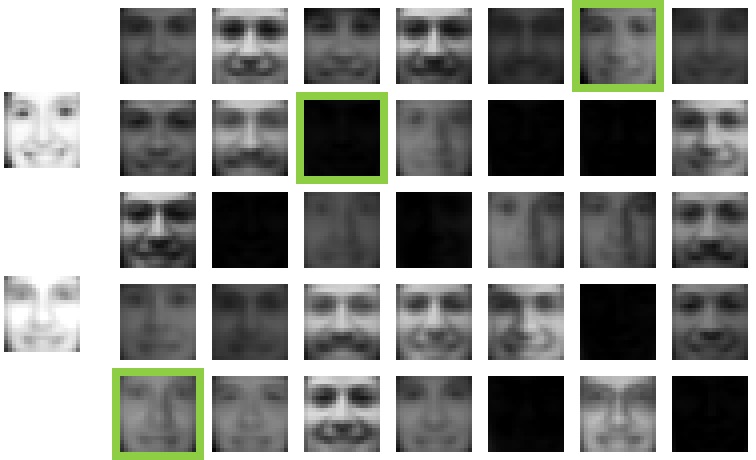}
  \caption{Face image reconstruction using features learned by NBMF.
  The five-by-seven matrix of images on the right shows the features that were learned.
  The two images on the left show the original image (top) and the reconstruction (bottom).
  The reconstruction is obtained by summing the features that are boxed in green.
  Note that although some of the features appear to be all black, they actually contain facial features are small in magnitude (black corresponds to 0, white corresponds to 1).}
  \label{fig:faces}
\end{figure}

\begin{figure}
  \includegraphics[width=\linewidth]{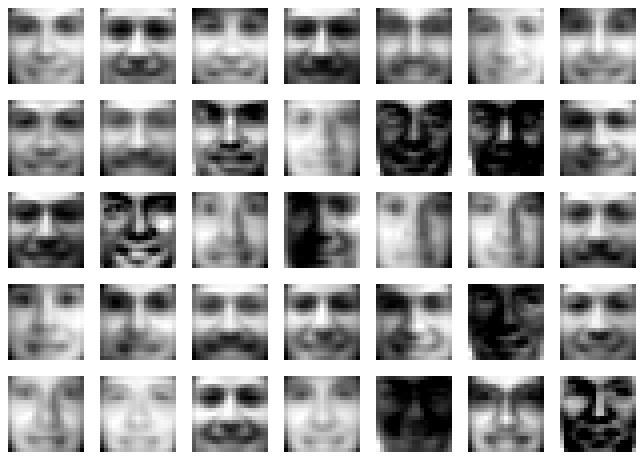}
  \caption{The same features as shown in figure \ref{fig:faces} are shown.
  Here they are rescaled to maximize contrast so that the darkest pixel is black and the brightest pixel is white.}
  \label{fig:contrastedfeatures}
\end{figure}

We analyzed the same set of 2,429 facial images that was previously analyzed to learn the parts of faces using nonnegative matrix factorization \cite{lee1999learning}.
Figure \ref{fig:faces} shows the features that were learned using algorithm \ref{alg:als} with 10,000 anneals per solve of equation \ref{eq:smallargmin} and how those features are used to reconstruct the image of a face.
Some of the features in figure \ref{fig:faces} may appear to be all black, but they actually contain subtle features such as a bright spot in the lower-left corner or a shiny cheek/nose (see figure \ref{fig:contrastedfeatures}).
Unlike NMF where the parts of faces are learned \cite{lee1999learning}, the features learned by NBMF are holistic.
One can view NMBF as being a method that is somewhere in between the NMF and vector quantization methods considered in \cite{lee1999learning}.
Like NMF, it imposes the nonnegativity constraints on $W$, but unlike NMF imposes binary constraints on $H$.
Vector quantization imposes binary constraints on $H$, but adds an additional constraint that each column of $H$ can only contain one nonzero entry.
This additional constraint causes vector quantization to learn holistic, prototypical faces.
NBMF appears to learn features that are holistic like the features that come out of vector quantization, but, unlike vector quantization, NBMF's features are not necessarily prototypical faces.
Many of them appear ghostly and the ones that are mostly black are even more subtle.
As in figure \ref{fig:faces}, these subtle and ghostly features can be combined to reproduce a face.

The NBMF method employed here has some advantages and disadvantages compared to NMF.
One advantage of NBMF is that $H$ is more sparse when NBMF is used than when NMF is used.
Analyzing this database of facial images using 35 features ($k=35$), the $H$ produced by NBMF is approximately 85\% sparse (i.e., 85\% of the elements of $H$ are zero) whereas the one produced by NMF is approximately 13\% sparse.
Further, the storage requirements for each component of $H$ are less for NBMF (1 bit) than NMF (e.g., 64 bit floating point numbers were used here).
A disadvantage is that $||V - WH||_F$ is larger for NBMF than NMF.
For this database of facial images, $||V - WH||_F$ using NMF was about 46\% of this norm when using NBMF.
In layman's terms, NMF had about half as much error as NBMF.
NMF has the additional advantage that $W$ is about 41\% sparse, whereas $W$ is dense for NBMF in the images analyzed here.

\begin{figure}
  \includegraphics[width=\textwidth]{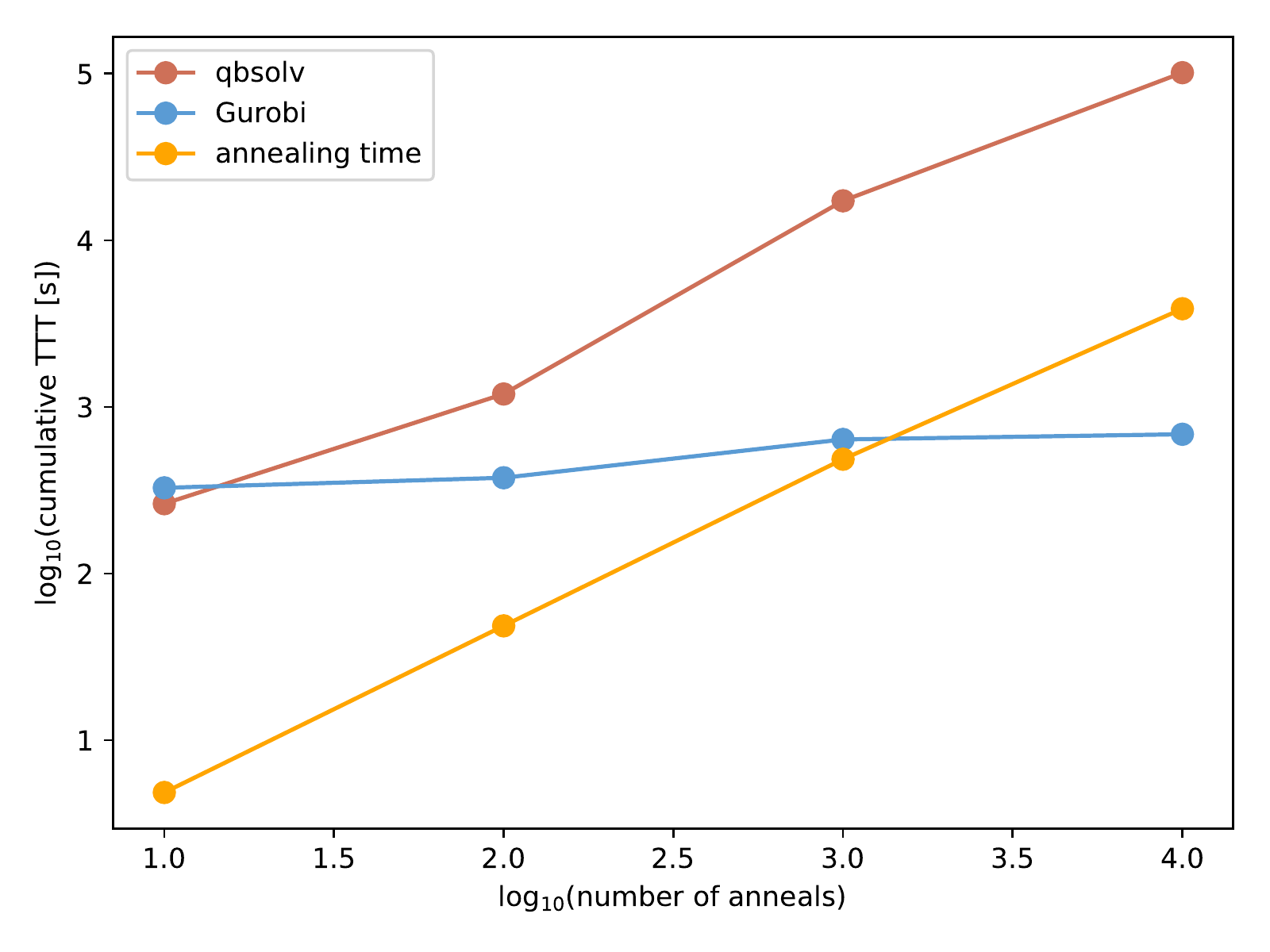}
  \caption{Cumulatve time-to-targets for qbsolv (red) and Gurobi (blue) as a function of the number of annealing cycles executed by the D-Wave.
  When the number of anneals is 10, 100, or 1,000, the cumulative time-to-targets for both qbsolv and Gurobi exceeds the cumulative annealing time (orange).
  When the number of anneals is 10,000, the cumulative time-to-targets for qbsolv exceeds the cumulative annealing time, but Gurobi's cumulative time-to-target is less than the cumulative annealing time.
  Note that in the 10, 100, and 1,000 anneal cases, 24,290 QUBOs were solved whereas only 19,432 QUBOs were solved in the 10,000 anneal cases.
  This was caused by earlier termination of the NBMF (algorithm \ref{alg:als}).}
  \label{fig:cttt}
\end{figure}

Figure \ref{fig:cttt} shows the results of the cumulative time-to-targets benchmark.
The cumulative time-to-targets for qbsolv always exceeds the cumulative annealing time by a factor of 20-50 depending on the number of anneals.
In the test with 10, 100, and 1,000 annealing cycles, Gurobi's cumulative time-to-targets exceeds the cumulative annealing time by factors of about 61, 7 and 1.2, respectively.
In the test with 10,000 anneals, Gurobi's cumulative time-to-targets was less than the cumulative annealing time by a factor of about 6.4.

\begin{figure}
  \includegraphics[width=\textwidth]{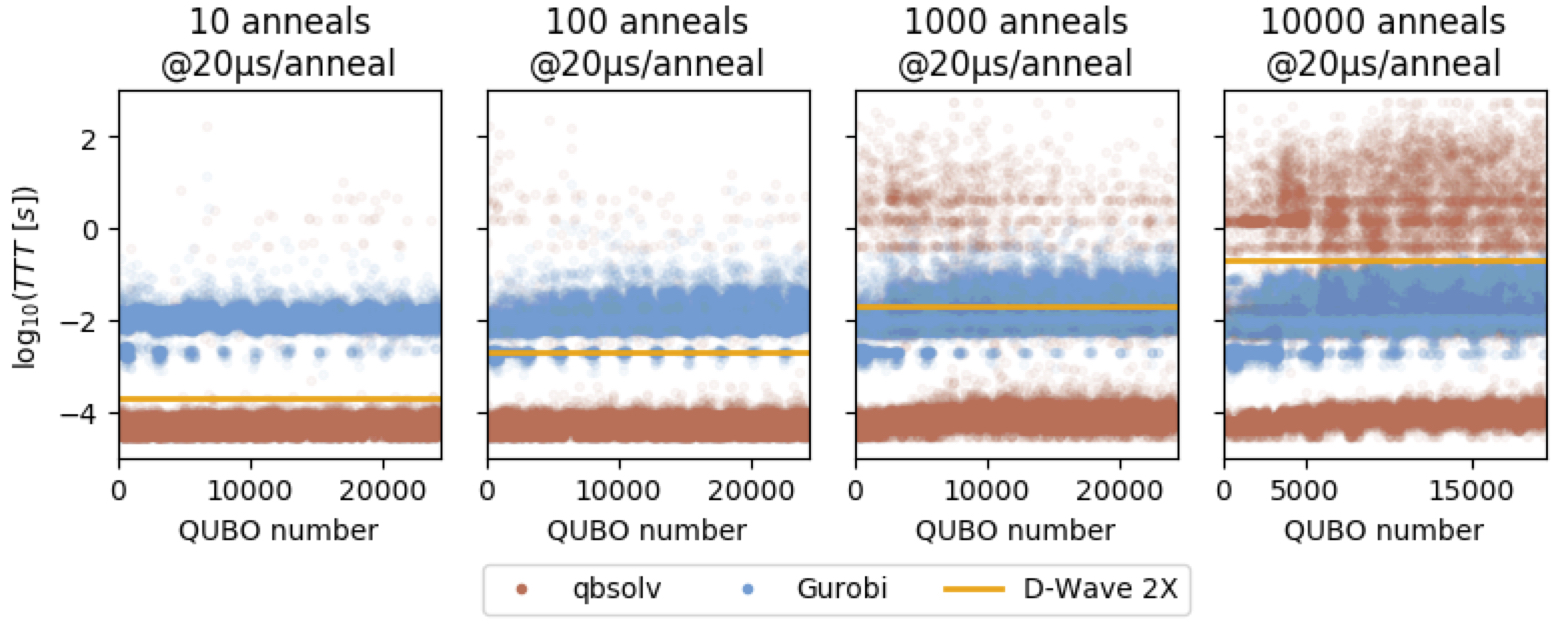}
  \caption{The time-to-target for each instance of equation \ref{eq:smallargmin} is shown for qbsolv (red dots), Gurobi (blue dots) in comparison to the annealing time used by the D-Wave (orange line) is shown.
  A different number of anneals was used in each execution of the NBMF (algorithm \ref{alg:als}) ranging from 10 to 10,000.}
  \label{fig:ttt}
\end{figure}

Gurobi and qbsolv show different performance trends in this benchmark that we explore in more detail with figure \ref{fig:ttt}.
Qbsolv's performance is characterized by frequently reaching the target before the annealing time for individual problems, but, when it fails to reach the target before the annealing time, it can take a comparatively long time to reach the target.
These problems where qbsolv takes a very long time to reach the target make up a large portion of the cumulative time-to-targets.
As the D-Wave takes more and more samples, the fraction of the problems where the time-to-target exceeds the annealing time increases, as can be seen from the increasing number of red dots above the orange line in figure \ref{fig:ttt}.
When 10 anneals are used, qbsolv's time-to-target exceeds the annealing time less than 1\% of the time, whereas when 10,000 anneals are used, qbsolv's time-to-target exceeds the annealing time in more than 28\% of the problems.
Gurobi rarely has problems for which it takes a very long time to reach the target set by the D-Wave.
In the more than 90,000 problems considered here, Gurobi took more than a second to reach the target set by D-Wave only 21 times with the maximum time being 13.6 seconds.
By contrast, qbsolv took more than a second to reach the target 5,509 times and hit the 10 minute maximum in 24 cases.
While Gurobi rarely takes a long time to reach the target set by the D-Wave, it also rarely solves the problems very quickly.
Gurobi reached the target set by the D-Wave in under a millisecond in only 57 cases, whereas qbsolv did this in almost 80,000 cases.
In short, Gurobi's performance is characterized by consistency and qbsolv's performance is characterized by solving many problems very quickly and some problems slowly.

\section{Discussion}\label{sec:discussion}
We have identified a performance regime (fast solutions that are good, but not necessarily optimal) where the D-Wave, at least by the benchmark used here, outperforms two state-of-the-art classical codes in a problem that is not custom-tailored to the D-Wave as other problems have been \cite{ronnow2014defining,denchev2016computational}.
This was demonstrated by the cases we studied with 10, 100, and 1,000 anneals where the cumulative time-to-targets for both Gurobi and qbsolv exceeded the annealing time used by the D-Wave.
We emphasize that these results do not demonstrate any sort of quantum supremacy.
In fact, they suggest a classical heuristic which would likely outperform the D-Wave: run qbsolv for a short period and if it fails to match the target, then switch to Gurobi.
This would leverage qbsolv's ability to frequently outperform the D-Wave, and Gurobi's ability to never lose too badly to the D-Wave.
Other benchmarks could also be used that might cast the D-Wave in a more negative light.
For example, allowing the D-Wave to set the targets in the cumulative time-to-targets benchmark might be an advantage, and reversing the roles (i.e., allowing Gurobi or qbsolv to set the targets in a fixed amount of computational time) might produce different results.

Given the remarkable performance improvements over many generations of classical microprocessors \cite{danowitz2012cpu} and the impressive algorithmic improvements in mixed-integer programming tools like Gurobi \cite{bixby2012brief} over the past several decades, it is surprising that D-Wave's third generation hardware and our straight-forward algorithm can be competitive at all.
In the series of four chips that D-Wave has released, the number of qubits has approximately doubled from one generation to the next while the number of couplers per qubit has remained essentially unchanged.
D-Wave's fifth generation chip is expected to at least double the number of couplers per qubit \cite{hilton2016systems,gibney2017d}.
If this comes to fruition, it would likely have a significant, positive impact on the performance of the D-Wave for the problems we consider here.
If the number of binary variables were fixed at 35, it would result in fewer physical qubits being used to represent each binary variable.
This means that, if the size of the problem is fixed, we would expect much better performance from D-Wave's fifth generation chip.
Increasing the number of couplers per qubit would also enable the possibility of going beyond the 35 binary variables considered here without using an excessive number of physical qubits to represent each binary variable.
That is, we would expect D-Wave's fifth generation chip to be capable of solving NBMF problems with more features than can be solved with the third generation chip that we have used here.
In addition to the improvements that are anticipated in future D-Wave hardware, there are techniques that could be utilized to potentially improve our algorithm.
Symmetries could be exploited (e.g., via spin reversal transformations or symmetry in the complete graph that we embed in D-Wave's hardware graph), the strength of chains arising from the embedding process could be optimized, hardware biases could be learned \cite{lokhov2016optimal} and exploited, and the embeddings could be improved by setting quadratic coefficients in the QUBO that are approximately zero to exactly zero.
Exploring and exploiting these techniques is beyond the scope of this manuscript, but we expect that some or all of them could provide significant algorithmic improvements.

The relatively short computational time (from $200\mu s$ up to $20,000\mu s$) performance regime where the D-Wave outperforms qbsolv and Gurobi in our benchmark could be important for big data problems.
For example, when learning the features of a large set of images (much larger than the 2,429 images considered here), only a small amount of computational time may be available to solve each problem given in equation \ref{eq:smallargmin}.
However, in order for the D-Wave's performance advantages in this regime to be beneficially leveraged, there must be significant performance improvements in the time it takes to get problems into and solutions out of the D-Wave.
At present, input and output is performed via the HTTPS internet protocol.
The performance of this bottleneck can clearly be improved.
Beyond this, there are other potential bottlenecks that could prevent the D-Wave from being more performant for these types of problems, such as the D-Wave's programming time (``qpu\_programming\_time'', as reported by D-Wave's software) which was typically is about 15 milliseconds in the problems analyzed here.
If this programming time remains constant as new D-Wave chips become available, then there would not be much advantage to solving problems where the total annealing time is much less than 15 milliseconds.
This could hinder the D-Wave's performance in the short computational time regime we have identified where it outperforms Gurobi and qbsolv in the cumulative time-to-targets benchmark.

In summary, we have demonstrated that this NBMF algorithm can leverage the D-Wave 2X as a key component in an unsupervised machine learning analysis.
Getting performance from the D-Wave 2X that is competitive with advanced classical tools on a real-world problem is a significant step forward on the journey towards practical quantum annealing.
While there is still much work to be done to make these quantum annealers of practical use for this type of problem, our performance results give a glimmer of hope that this may someday be the case.

\end{document}